\documentclass[letterpaper, 10 pt, conference]{ieeeconf}

\usepackage[disable]{todonotes}

\usepackage{xcolor}
\newcommand{\subparagraph}{}
\usepackage{titlesec}

\reversemarginpar 

\usepackage{siunitx}

\usepackage{dsfont}
\usepackage{amsmath}
\usepackage{bm}
\usepackage{bbold} 
\usepackage[font=small]{subcaption,caption} 

\usepackage{multirow} 
\usepackage{makecell} 

\usepackage[utf8]{inputenc}
\usepackage[T1]{fontenc}

\newcommand{\italic}[1]{\textit{#1}}
\renewcommand{\b}[1]{\bm{#1}}

\newcommand{\fxut}{\b{f}_{\b{xu}_t}}
\newcommand{\fxutreg}{\b{f}^{'}_{\b{xu}_t}}

\newcommand{\fct}{\b{f}_{\b{c}_t}}
\newcommand{\Ft}{\b{F}_t}

\newcommand{\dkl}{D_{\mathrm{KL}}}

\renewcommand{\xi}{\b{x}_{i}}

\newcommand{\xt}{\b{x}_{t}}

\newcommand{\xtt}{\b{x}_{t + 1}}
\newcommand{\ut}{\b{u}_{t}}

\newcommand{\ph}{{\phi_h}}
\newcommand{\pk}{{\phi_k}}

\newtheorem{remark}{Remark}

\usepackage[noabbrev, capitalise]{cleveref}

\IEEEoverridecommandlockouts                              

\usepackage{amssymb}  

\renewcommand{\baselinestretch}{.97}

\title{
\LARGE\bf Learning a Structured Neural Network Policy for a Hopping Task
}

\author{Julian Viereck$^{1, 2}$, Jules Kozolinsky$^{3}$, Alexander Herzog$^{4}$, Ludovic Righetti$^{1, 2}$
\thanks{This research was supported by New York University, the Max-Planck
Society, the Max Planck ETH center for learning systems and the European Union’s Horizon 2020 research and innovation
programme (grant agreement No 780684 and European Research Council's
grant No 637935).}
\thanks{$^{1}$Tandon School of Engineering, New York University, USA}%
\thanks{$^{2}$Max Planck Institute for Intelligent Systems,
  Germany {\tt\footnotesize firstname.lastname@tuebingen.mpg.de}}%
\thanks{$^{3}$Ecole Normale Supérieure
  Paris-Saclay
  France {\tt\footnotesize jules.kozolinsky@ens-cachan.fr}}%
\thanks{$^{4}$X, Mountain View, California, USA {\tt\footnotesize alexherzog@x.team}}%
}

\begin{document}

\maketitle

\begin{abstract}

In this work we present a method for learning a reactive policy for a simple dynamic locomotion task involving hard impact and switching contacts where we assume the contact location and contact timing to be unknown. To learn such a policy, we use optimal control to optimize a local controller for a fixed environment and contacts. We learn the contact-rich dynamics for our underactuated systems along these trajectories in a sample efficient manner. We use the optimized policies to learn the reactive policy in form of a neural network. Using a new neural network architecture, we are able to preserve more information from the local policy and make its output interpretable in the sense that its output in terms of desired trajectories, feedforward commands and gains can be interpreted. Extensive simulations demonstrate the robustness of the approach to changing environments, outperforming a model-free gradient policy based methods on the same tasks in simulation. Finally, we show that the learned policy can be robustly transferred on a real robot.

\end{abstract}


\section{Introduction}
%
For robots to be fully autonomous it is important that control policies deal with fast changes in the environment. This is particularly true for locomotion tasks involving dynamics and unknown contact situations.
Over the past years, trajectory optimization methods have been successfully applied to the control of complex locomotion tasks \cite{PosaT12,tassa2014control,Herzog-2016b,mastalli2016hierarchical}. These methods tend to require significant computation time and usually assume a static environment. A way to run the computation in real time and react to changes in the environment is by using model predictive control \cite{erez2013integrated,koenemann2015whole}. Lately, these techniques were verified on real hardware and have been proven robust for unspecified contact timing and location \cite{neunert2017whole}. However, they are still to slow to handle an unexpected disturbance at contact during fast motions such as jumping. Moreover, contact switching is known to be an issue for most trajectory optimization methods~\cite{posa2016optimization}.
\begin{figure}[th]
    \centering
       \includegraphics[width=0.88\linewidth]{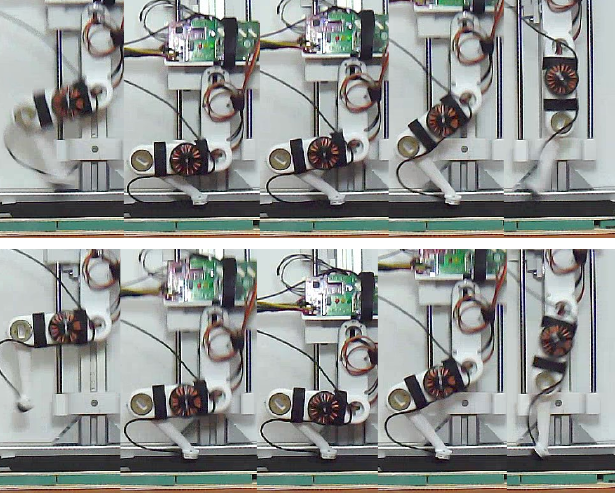}
    \caption{One jump sequence on the real robot running the feedback network policy (120ms between each frame).}
    \label{fig:fig:jump_traj_single}
    \vspace{-0.5cm}
\end{figure}

To alleviate the computational demand of online optimization methods, approaches that directly learn policies have been investigated.
For example, trajectories and neural networks are optimized together in~\cite{mordatch2014combining}. In Guided Policy Search (GPS)~\cite{levine2013guided}, trajectories are optimized using iterative Linear Quadratic Regulator (iLQR)~\cite{li2004iterative} with an augmented cost to enforce the neural network and trajectory optimization to converge to similar results.
The advantage of such methods is that it is not necessary to optimize online new control laws as the environment changes as long as the learned policy is able to generalize to the new environment.
However, to our best knowledge, these approaches have never been demonstrated on a real robot with under-actuated dynamics for unstable tasks with non-negligible impact dynamics.

Neural network policies have also been successfully trained using "model-free" reinforcement learning. While there has been significant progress in playing games~\cite{mnih2015human,silver2016mastering,silver2017mastering} and for the simulation of animated characters~\cite{heess2017emergence}, the required amount of samples to learn the policies is prohibitively high and it remains to be shown that such policies can be transferred on real hardware.

The advantage of learning local dynamic models to compute locally optimal control laws has been demonstrated a long time ago in~\cite{schaal94}. More recently, this has been exploited further in the context of the GPS framework~\cite{levine2015learning} and in~\cite{meier16} for inverse dynamics learning.
However, as we discuss in this paper, such approaches lead to ill-conditioned dynamics in the case of under-actuated systems and limiting their use to compute locally optimal control laws.
Learned dynamics models allows us to use sampling-based approaches with efficient optimal control methods such as iLQR when using a simulator. This proves useful when dealing with complex, discontinuous dynamics for which computing the gradient from the dynamics model can be a numerically expensive and ill-conditioned process.

In this paper, we aim to learn a reactive feedback policy
for tasks involving fast contact dynamics and under-actuation, that remains robust to changes in the environment and
in particular to contact changes.
In particular, we show 1) how local dynamic models can be learned for systems with under-actuated dynamics, 2) how iterative LQR with an adaptive receding horizon can be leveraged to learn unstable tasks and 3) how we can learn a policy neural network that is amenable to analysis by preserving the feedback/feedforward control structure. We call the last point interpretability of the network policy which is interesting to us as it opens a path to use optimal control techniques to analyse the network's output and thereby bridge the gap between classical control and (modern) learning techniques.
We conduct extensive simulation and real robot experiments on a single-legged robot performing a dynamic hopping task which exhibits under-actuated and discontinuous dynamics.
We show that our approach can generalize to unseen terrains. We compare the learned policy with more traditional torque output networks and show that it retains similar performance while allowing for an easier to analyze policy. We also show that it tends to perform better than
a model-free policy gradient reinforcement learning algorithm with significantly smaller sampling complexity in simulation. Finally, we demonstrate the capabilities of the trained policies on a real robot.
\section{Dynamics Learning}
\label{sec:dyn_learning}
%
%
\subsection{Basic notation}
We denote the measured states of a robot at time $t$ by $\xt$. The actions sent to the robot are denoted by $\ut$. We work in a time discretized system. We write the cost at time $t$ as $\b{\ell}(\xt, \ut)$. For a trajectory $\b{\tau} = \{\b{x}_1, \b{u}_1, ..., \b{x}_T, \b{u}_T\}$ the trajectory cost is given by the sum over the costs at every timestep. We use stochastic control policies, which we model as Gaussian distributions $p(\ut|\xt) = \mathcal{N}(\b{\mu}_t, \b{\Sigma}_t)$. For modeling the dynamics, we assume a Gaussian model, where the mean is an affine function of the robot's current state and action:
\begin{align}
\label{eq:def_dynamics}
p(\xtt | \xt, \ut) = \mathcal{N}\left(\fxut \begin{bmatrix} \xt \\ \ut \end{bmatrix} + \fct, \Ft \right),
\end{align}
where $\Ft$ and $\fxut$ are matrices and $\fct$ a constant offset.
\subsection{Learning the dynamics model}
The first stage of the algorithm is to learn local dynamic models with hard contact switches. These models are
then used with an iterative LQR algorithm to obtain locally optimal policies.
Such local models were successfully learned for fully actuated robots to achieve tasks that did not involve
hard impact dynamics (e.g. \cite{levine2015learning}) by fitting affine dynamics models along recorded trajectories.
A dynamics model along a trajectory is learned by running a stochastic policy N times. Let $\b{s}^i_t$ be the transition at time $t$ for the $i$-th rollout, with $\b{s}^i_t=(\xt, \ut, \xtt)$. At every timestep, a Gaussian distribution is fitted assuming the transition model $p(\xt, \ut, \xtt) = \mathcal{N}(\b{\mu}_t, \b{\Sigma}_t)$. Computing the conditional probability distribution $p(\xtt | \xt, \ut)$ in closed form~~\cite[Chapter~2.3]{Bishop:2006:PRM:1162264} and comparing terms to Eq. \eqref{eq:def_dynamics}, one arrives at
\begin{align}
\fxut   &= \b{\Sigma}_{\xtt, \b{xu}_t} \b{\Sigma}^{-1}_{\b{xu}_{t}, \b{xu}_{t}} \label{eq:dyn_fxut}, \\
\fct    &= \b{\mu}_{\xtt} - \fxut \b{\mu}_{\b{xu}_t} \label{eq:dyn_fct}, \\
\Ft       &= \label{eq:prob_dyn_covar}
        \b{\Sigma}_{{\xtt},{\xtt}} - \b{\Sigma}_{{\xtt}, \b{xu}_{t}} \b{\Sigma}_{\b{xu}_{t}, \b{xu}_{t}}^{-1} \b{\Sigma_{\b{xu}_{t},\xtt}},
\end{align}
where the indices correspond to accessing the subparts of the vectors and matrices. To reduce the required sample count N, in~\cite{levine2015learning} a Gaussian Mixture Model (GMM) is fitted to all transitions $s^i_t$ along the N trajectories and also includes trajectories from previous iterations. The GMM is then used as a prior when estimating $p(\xt, \ut, \xtt)$.

Learning a dynamics model this way is very attractive as it is sample efficient and computationally simple. However, in the case
of underactuated dynamics, which correspond to the class of problems studied in this paper, such a
local regression approach does not work properly.
In our experiments, the dynamics model fitting always led to physically implausible,
ill-conditioned, $\fxut$ matrices in Eq. \eqref{eq:dyn_fxut},  independently of the number
of samples.
These issues stem from the fact that the floating-base position and velocity are not directly controllable when the robot is in free air. Current state of the art approaches \cite{levine2015learning} to learn local dynamic models are ill posed for these type of systems and lead to numerical issues that prevent their use for local policy optimization.

We address this issue by regularizing the dynamics matrix: we introduce a prior on the $\fxut$ dynamics matrix that assumes the change in state is close to identity.
To introduce a zero-mean prior, we define a shifted dynamics matrix $\fxutreg = \fxut - (\mathbb{1}~\huge0)$ and then assume a prior as $\fxutreg \sim \mathcal{N}(0, \lambda^{-1} \mathbb{1})$. Similarly, we change the transition points to $\b{s}^{i'}_t=(\xt, \ut, \xtt - \xt)$. Using results from~\cite[Section 3.4]{williams1998prediction}, the dynamics parameters are now estimated by:
\begin{align}
\b{f}_{\b{xu}_t, \textrm{ridge}}  =& \b{\Sigma}_{\xtt, \b{xu}_t} (\lambda \mathbb{1} + \b{\Sigma}_{\b{xu}_{t}, \b{xu}_{t}})^{-1} \label{eq:dyn_fxut_ridge} + (\mathbb{1}~\huge0), \\
\b{F}_{t, \textrm{ridge}}     = &\label{eq:prob_dyn_covar_ridge}
        \b{\Sigma}_{{\xtt},{\xtt}} - \{\b{\Sigma}_{{\xtt}, \b{xu}_{t}}  \\& \nonumber
        \qquad(\lambda \mathbb{1} + \b{\Sigma}_{\b{xu}_{t}, \b{xu}_{t}})^{-1} \b{\Sigma}_{\b{xu}_{t},\xtt} \}.
\end{align}
In our experiments, the regularization term remains small ($\lambda \sim 10^{-2}$) and yet is
sufficient to have stable learning.
\subsection{Handling torque limits}
We consider very dynamic tasks where output torques are often close to the robot's actuation limits.
This can cause issues when learning the dynamics as we might attempt to sample torque commands from $p(\ut|\xt)$ in~Eq. \eqref{eq:ilqr_policy} exceeding these limits. This can create a bias in the estimated dynamics. We eliminate this issue by using lower torque limit: we allow a maximum torque 90\% of the maximum torque. The remaining 10\% are then available for sampling when learning the dynamics.
\begingroup
\titlespacing{\section}{0pt}{1.0cm}{0.2\baselineskip}
\section{iLQR policy optimization}
\endgroup
\label{sec:ilq_policy_optimization}
To optimize locally optimal feedback policies around a trajectory, we follow the work by~\cite{levine2015learning}
and use iterative LQR, as introduced in \cite{li2004iterative,Sideris:2005hp}.
%
%
%
iLQR is a single shooting method that computes approximations to the derivatives of the value~$\b{V}$ and $\b{Q}$ function to solve the backward Riccati equations and use the full model dynamics during the shooting phase. We denote with $\b{A}_{\b{xu}t}$ the Jacobian of $\b{A}$ and with $\b{A}_{\b{xu},\b{xu}t}$ its Hessian at time $t$ with respect to the concatenated state and action vector $\b{xu}$. The backwards pass recursion to compute the value and Q functions reads:

\begin{align}
\b{Q}_{\b{xu}t} &= \label{eq:q_xut}
  \b{\ell}_{\b{xu}t} + \fxut^\mathsf{T} \b{V}_{\b{x}{t+1}} +
  \fxut^\mathsf{T} \b{V}_{\b{x}, \b{x}{t+1}} \fct \\
\b{Q}_{\b{xu}, \b{xu}t} &= \label{eq:q_xuxut}
  \b{\ell}_{\b{xu}, \b{xu}t} +
  \fxut^\mathsf{T} \b{V}_{\b{x},\b{x}{t+1}}\fxut \\
\b{V}_{\b{x}t} &= \label{eq:v_xut}
  \b{Q}_{\b{x}t} - \b{Q}^\mathsf{T}_{\b{u},\b{x}t} \b{Q}^{-1}_{\b{u},\b{u}t} \b{Q}_{\b{u}t} \\
\b{V}_{\b{x},\b{x}t} &= \label{eq:v_xuxut}
  \b{Q}_{\b{x},\b{x}t} - \b{Q}^\mathsf{T}_{\b{u},\b{x}t} \b{Q}^{-1}_{\b{u},\b{u}t} \b{Q}_{\b{u},\b{x}t},
\end{align}

with the derivatives of $\b{V}$ for $t > T$ set to zero. In our problems, it is important to take into account the actuation
limits of the robot and we use the method described in ~\cite{tassa2014control} to compute the optimal control command.
%
%
%
After the backwards pass, a new policy is synthesized in the shooting phase as
\begin{align}
  \label{eq:ilqr_policy}
  p(\ut|\xt) &= \mathcal{N}(\b{k}_t + \b{K}_t (\xt - \hat{\b{x}}_t), \b{\Sigma}_t)\\
  \b{k}_t &= \hat{\ut} + \alpha \delta \ut,
\end{align}
where $\hat{\ut}$ is the action along the provided trajectory, $\delta \ut$ is the computed change of action to minimize the local cost and $\hat{\b{x}}_t$ is the new desired trajectory. We denote the new feedforward action of this policy by $\b{k}_t$. Following~\cite{tassa2014control}, we zero entries in the feedback matrix $\b{K}_t$ that correspond to actions exceeding torque limits. The parameter $\alpha \in [0, 1]$ is used for linesearch with the intention to find the update step size corresponding to the maximum cost reduction.
\subsection{Parameter choices for unstable tasks}
Instead of the linesearch, we use a trust region for the control update, as in~\cite{levine2015learning}.
We set $\alpha = 1$ and enforce a constraint between the new $p_\text{new}$ and previous $p_\text{old}$ optimized policy. The constraint is given by $\dkl(p_\text{old}(\ut|\xt)|p_\text{new}(\ut|\xt)) \leq \epsilon$, where $\dkl(p|q)$ denotes the Kullback-Leibler-Divergence. To enforce the constraint, the initial loss is augmented to be
\begin{align}
\tilde{\ell}(\xt, \ut) = \frac{1}{\eta}\ell(\xt, \ut) - \log p_\text{old}(\ut|\xt).
\end{align}
We perform a line search on $\eta$ in log space until the KL constraint is satisfied.
However, we made two important modifications compared to previous work.
First, we set $\epsilon = 0.5$ in our experiments. While it is also possible to use a heuristic
approach to adapt it during the optimization as in~\cite{levine2016end}, we noticed that
using a fixed $\epsilon$ turned out to give better
convergence results than using the heuristic with the benefit of a simpler algorithm.

Second, we restrict the policy's covariance to values typically around $\b{\Sigma}_t=0.01\mathds{1}$, instead of
using $\b{\Sigma}_t=\b{Q}^{-1}_{\b{u},\b{u}t},$ for the feedback policy, as was done in previous work \cite{levine2015learning}.
Indeed, we noticed that such a choice does not work properly in the case of impacts. In our experiments, we observed large eigenvalues of $\b{\Sigma}_t$ which  in turn created unstable robot behavior.
This effect is especially important when the robot is about to touch the ground. In this case, large uncertainties in the motion causes the robot to fall easily, preventing the optimization to find any stable gait.
%

%
\subsection{Adaptive receding horizon length}
%
%
We start the iLQR optimization procedure from a randomly initialized policy. As the task we study is intrinsically unstable, the initial policy is likely to make the robot fall. This can be a problem since the iLQR iteration is likely to not converge.
To tackle this issue, we propose to automatically increase the receding horizon length over the iterations instead of optimizing a task with a fixed horizon length.
It allows us to find optimal policies quickly, as the backward Riccati pass is able to better propagate the value function over
small horizon at the beginning when the trajectory is unstable.
Once the hopping motion is stable enough, the trajectory gets extended at a maximum of 50 ms per optimization iteration. In this context, stable enough means the robots base is high enough above the ground. The gradual extension of the optimization horizon allowed the algorithm to quickly find longer jumping trajectories.
%
\subsection{Cost function design}
\begin{figure}
  \centering
    \vspace{0.2cm}
    \includegraphics[scale=0.65]{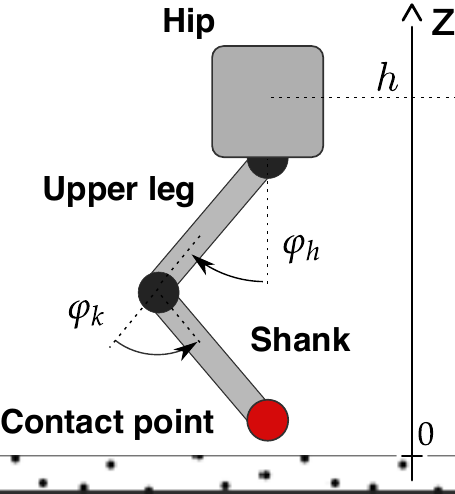}
    \setlength{\tabcolsep}{0.2em} 
    \renewcommand{\arraystretch}{1.2}
    \small
    \begin{tabular}[b]{l|l|l}
        Property & Simulation & Real\\ \hline
        $m_\text{Hip}$ & \SI{0.15}{\kilogram} & \SI{0.265}{\kilogram}\\ \hline
        $m_\text{Upper leg}$ & \SI{0.1}{\kilogram} & \SI{0.145}{\kilogram}\\ \hline
        $m_\text{Shank}$ & \SI{0.01}{\kilogram} & \SI{0.025}{\kilogram}\\ \hline
        \makecell[tl]{Leg length} & \SI{0.2}{\meter} & \SI{0.2}{\meter} \\ \hline
        \makecell[tl]{Max torque} & \SI{1.3}{\newton\meter} & \SI{0.675}{\newton\meter} \\ \hline
        \makecell[tl]{Joint damping} & -0.05 $\dot{\phi}_{h, k}$ & unknown  \\
    \end{tabular}
%
  \caption{Experimental platform: a single leg hopper with base constrained to move along the vertical axis.}
  \label{fig:hopper_setup}
  \vspace{-0.5cm}
\end{figure}

Our cost function is constituted for two terms that describe
the different behaviors expected during different dynamic modes when performing a hopping motion.
We use the term $\ell_\textrm{jump}$ for situations when the robot is close to the ground to makes the robot jump. When the robot is in the air, we defined $\ell_\textrm{land}$ to put the leg in a useful position:
\begin{align}
    \label{eq:cost_hopper_jump}
    \ell_\textrm{jump}(\xt, \ut) = &
          0.2\, \left(\ph - 0.0\right)^2 + 0.2\, \left(\pk - 0.0\right)^2 +\\&  0.001\, \left({u_h}^2 + {u_k}^2\right) \nonumber \\
    \ell_\textrm{land}(\xt, \ut) = &
        0.2\,\left(\ph - 1.3\right)^2 + 0.2\, \left(\pk - 2.3\right)^2 + \\ & 0.001\, \left({u_h}^2 + {u_k}^2\right), \nonumber
\end{align}
where $\phi_h$ and $\phi_k$ denote the hip and knee angle and $u_h$ as well as $u_k$ the action applied at each of these joints (see Eq.\eqref{fig:hopper_setup}). We also tried to optimize a cost that reflects the desired goal more directly (like maximizing the upwards hip velocity). However, the above costs led to better and more reproducible results, even when using
ground truth dynamics.
During the backward pass of iLQR, the value and Q function derivatives are reset whenever the cost switches
to prevent the propagation of the previous cost in the policy.
%
%
\section{Reactive neural network policy}
\label{sec:reactive_neural_network_policy}
The optimized iLQR policies are indexed by time, which is an issue in the case of contact dynamics as time
indexing assumes fixed timing for contact switching. Thus, if contact happens earlier or later than the policy prediction, it is likely lead to a fall.
It is therefore desirable to learn a feedback policy that is independent of time.
\subsection{Feedback network and torque network policies}
\begin{figure}[t]
  \centering
  \vspace{0.2cm}
  \includegraphics[width=.9\linewidth]{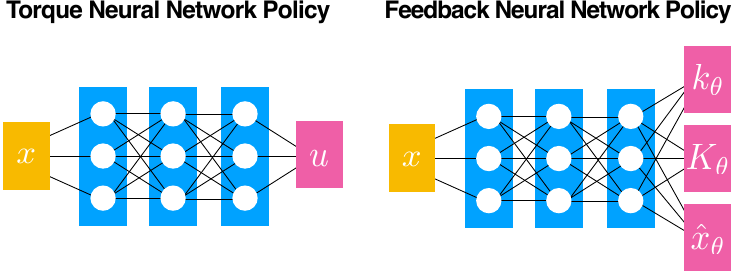}
  \caption{Neural network architectures used to learn policies.
  }
  \label{fig:network_overview}
  \vspace{-0.5cm}
\end{figure}

We train two different kinds of neural networks, a \italic{feedback neural network} and a \italic{torque neural network}, see Figure \ref{fig:network_overview}.
The torque neural network policy directly maps the input state $\b{x}$ to the action mean $\b{g}^\text{torque}_\theta(\b{x})$. This architecture has the advantage to directly map inputs to torques and has been successfully used in several previous works~\cite{levine2015learning, levine2016end, schulman2015trust}. However, it is hard to understand what the actual policy does, as we do not have access to an informative policy structure.
On the other hand, the feedback neural network policy maps the input state $\b{x}$ to a feedforward $\b{k}_\theta(\b{x})$ vector, feedback matrix $\b{K}_\theta(\b{x})$ and nominal position $\hat{\b{x}}_\theta(\b{x})$. Similar to the iLQR policy, the final action mean is then computed as $\b{g}^\text{feedback}_\theta = \b{k}_\theta(\b{x}) + \b{K}_\theta(\b{x}) (\b{x} - \hat{\b{x}}_\theta(\b{x}))$. This network architecture allows us a direct interpretation of the network output in terms of feedforward and feedback control paths which have very different physical interpretations. We discuss such interpretations in the experimental section.

For both kinds of networks, an action is computed as $\b{u}=\mathcal{N}(\b{g_\theta},\b{\Sigma}_\theta)$ and we use ELUs as an activation function. We trained networks with different parameters (number of layers, number of hidden neurons). Eventually, we choose a configuration with small number of parameters and that performs in the top 10\% on the training data over all trained networks. We use three fully connected layers with 10 hidden neurons for the torque neural network.
In the feedback neural network, the input state $\b{x}$ is passed through three fully connected layers with 10 hidden neurons. The last neurons are fully connected and output feedforward torques, feedback gains and desired joint positions.
%
%
%

\subsection{Training procedure}
%
The network policies are trained with two iLQR policies for trajectories of length \SI{1}{\second} using learned dynamics and the method described in Section \ref{sec:ilq_policy_optimization}. For each optimization iteration, we sample five trajectories to estimate the dynamics and use $k=5$ Gaussian mixture components for the GMM prior. Each policy is optimized for 50 iLQR iterations. The iLQR policies have different initial positions given by
\begin{align}
x^1_0 & =  \{h, \dot{h}, \phi_h, \dot{\phi_h}, \phi_k, \dot{\phi_k}, c\}\\
  & = \{0.2, 0.0, 1.3, 0.0, 2.3, 0.0, 0.0\} \nonumber \\
x^2_0 & = \{0.3, 0.0, 1.0, 0.0, 1.55, 0.0, 0.0\}.
\end{align}
we generate the training data for the network policies using these policies. We increase the covariance of the policies in Eq. \eqref{eq:ilqr_policy} in the range of $\gamma \in [0., 0.30]$ as $\b{\Sigma}_t=\gamma \mathbb{1}$ in steps of $0.01$. For a set $\b{\Sigma}_t$, we sample five trajectories for each policy. It means that we need in total 250 episodes to compute an iLQR policy. Beside the trajectories, we store for each state $\b{x}^i_t$ the commanded action by the iLQR policy without noise $\b{g}^i_t$, as well as $\b{k}^i_t$, $\b{K}^i_t$ and $\hat{\b{x}}^i_t$.

With this training data of $M$ points $\b{x}^i_t$ and associated data, we optimize different losses for each kind of network in a supervised fashion. The losses are defined as:
\begin{align}
\label{eq:loss_nn_feedback}
\mathcal{L}^{\text{torque}}_\theta = \frac{1}{M} \sum^{M}_{i=1} &\|\b{g}^i_t - \b{g}^\text{torque}_\theta(\b{x})\|^2_2 \\
\mathcal{L}^{\text{feedback}}_\theta = \frac{1}{M} \sum^{M}_{i=1} \{&\|\b{k}^i_t - \b{k}^\text{fb}_\theta(\b{x})\|^2_2 + \|\b{K}^i_t - \b{K}^\text{fb}_\theta(\b{x})\|^2_2  \nonumber + \\
& \|\hat{\b{x}}^i_t - \hat{\b{x}}^\text{fb}_\theta(\b{x})\|^2_2 \} \nonumber,
\end{align}
where $\|\cdot\|_2$ denotes the Euclidean norm. We compute the norm for the feedback matrices by reshaping the matrices into a 1D vector. We optimize the losses using the Adam optimizer and stochastic sampled mini-batches of size 1024. We found the best learning rate for the torque network policy to be $0.0001$ and for the feedback network policy to be $0.001$. We optimized the networks for 20'000 batches each.
\section{Experimental setup}
\label{sec:experiments}
In this section we describe the experimental setup used in this contribution. We also describe the setup
for the Proximal Policy Optimization \cite{schulman2017proximal}
that we use as comparison with a state of the art model free reinforcement
learning algorithm. In this paper, we only
consider simulations similar to the real experimental platform.
%
%
%
%
%
\subsection{Performance metric}
To evaluate the performance of a policy, we define the ``Positive Hip Velocity Squared'' (PHVS) performance metric. This metric is computed as the mean of the upwards hip velocity squared, defined as:
\begin{align}
  \label{eq:policy_perf_metric}
  P(\tau) = \frac{1}{T} \sum_{t=1}^{T} \max(0, \dot{h}_t)^2.
\end{align}
This metric favors policies that make the hopper robot take off from the ground with a high velocity. Comparing many sample trajectories, this captures the notion of a decent performing hopping motion quite well.
\subsection{Experimental platform}
\label{sec:exp_setup}
The experimental platform is a custom-designed hopper robot. \cref{fig:hopper_setup} gives
an overview of the platform and important parameters.
The robot has two joints actuated by torque controlled brushless motors (T-motor Antigravity 4004 KV300), which allows for fast yet compliant motions. Each joint has an Avago Encoder with 5000 counts per resolution. The robot's unactuated base is constrained to move only in the vertical direction. An OptoForce OMD-20-SE-40N force sensor is mounted on the foot.

The robot is simulated using the \italic{Rigid Body Dynamics Library}~\cite{RBDL} implementation
of the Articulated Body Algorithm. We use a total momentum preserving contact model following~\cite[Section 3.4]{westervelt2007feedback}.


\begin{remark}
This contact model has been extensively used in the locomotion literature.
It gives a realistic, hard contact with discontinuity at impact and does not require to tune any parameters. Note that since there is only one contact point, the model is equivalent to a complementary constraint contact model without sliding friction since there is not force allocation problem. Due to the contact model, the dynamics is not
differentiable at impact which can therefore be an issue for gradient-based optimization methods. As we show in the following, our sampling-based learning approach allows us to optimize the robot motion. While the contact model does not capture sliding friction effects,
it was sufficient to learn a control policy robust on a real robot facing such effects.
\end{remark}

The control policies run in realtime at \SI{100}{\hertz} both in simulation and on the real robot. Besides damping at the joints, we assume no other source of friction in the simulation.
The measured robot state is its base-height above the ground, the hip and knee angle as well as velocities of these quantities.
For all the experiments, we use a binary contact state $c$, which is set to 1 in case the contact forces exceed a cutoff value.
Overall, the 7 dimensional state space is defined as ${\b{x} = (h, \dot{h}, \phi_h, \dot{\phi_h}, \phi_k, \dot{\phi_k}, c)}$.
%
\subsection{Comparison with Proximal Policy Optimization}
We compare our results with a policy trained using Proximal Policy Optimization (PPO)~\cite{schulman2017proximal}. We call this method PPO Network Policy. PPO is a state of the art model-free, reinforcement learning based method. The policy network is made of two hidden layers with 16 neurons each and ELU activation and is trained on four cores in parallel for 21,000 episodes each. We adjust the learning rate to 0.003 and follow otherwise the PPO1 baselines~\cite{baselines} implementation and it takes approximately 5000 episodes for PPO to converge. We also considered using an architecture similar to the feedback network. However, in our experiments we were not able to optimize such a network using PPO. This might be as for the feedback network, the policy needs to learn the feedforward, feedback and desired position, which are intertwined for the final control. Learning these three quantities from the reward signal alone is a more complicated task then learning to predict the control directly.
The learning scenario is the same as for the iLQR optimization and the system is reset to the $x^1_0$ and $x^2_0$ start position in an alternating fashion. An episode is terminated when the hip is less than \SI{0.1}{\meter} from the ground or the maximum of $T=100$ timesteps is reached. While there is a wide variety of techniques (dynamic perturbations,
sensor noise, etc) to make the PPO policy more robust, we omitted these to make the training of the PPO closer to the one for the local policies and thereby the results of the final network policies more comparable. We were not able to use the same cost function, as when using the two terms
$\ell_\textrm{jump}$ and $\ell_\textrm{land}$ the PPO algorithm learns a policy that remains and optimizes the $\ell_\textrm{land}$ cost all the time.
This result can be explained as switching to the $\ell_\textrm{jump}$ cost increases the overall cost locally. Therefore, the PPO algorithm learns to not jump. In order to still allow comparison with our approach,
we use a reward $r_t$ at each timestep $t$ that motivates upwards motion:
\begin{align}
  r_t =& -(\ph - 1.3)^2 - (\pk - 2.3)^2 + \max(0, \dot{h}_t)^2. \nonumber
\end{align}
The first two terms of this reward have an analogous function as the terms in $\ell_\textrm{land}$ to put the leg into a reasonable position for landing. These two terms are set to zero if the hip velocity is larger than \SI{0.2}{\meter/\second}. The third term motivates the robot to jump. When the episode ends at timestep $t$ we emit the penalty $ -(T - t)$ to motivate the optimization in finding trajectories lasting for the full $T$ timesteps.

\section{Results}
\begin{figure}
    \centering
    \includegraphics[width=0.9\linewidth,clip,trim={0 10 0 10}]{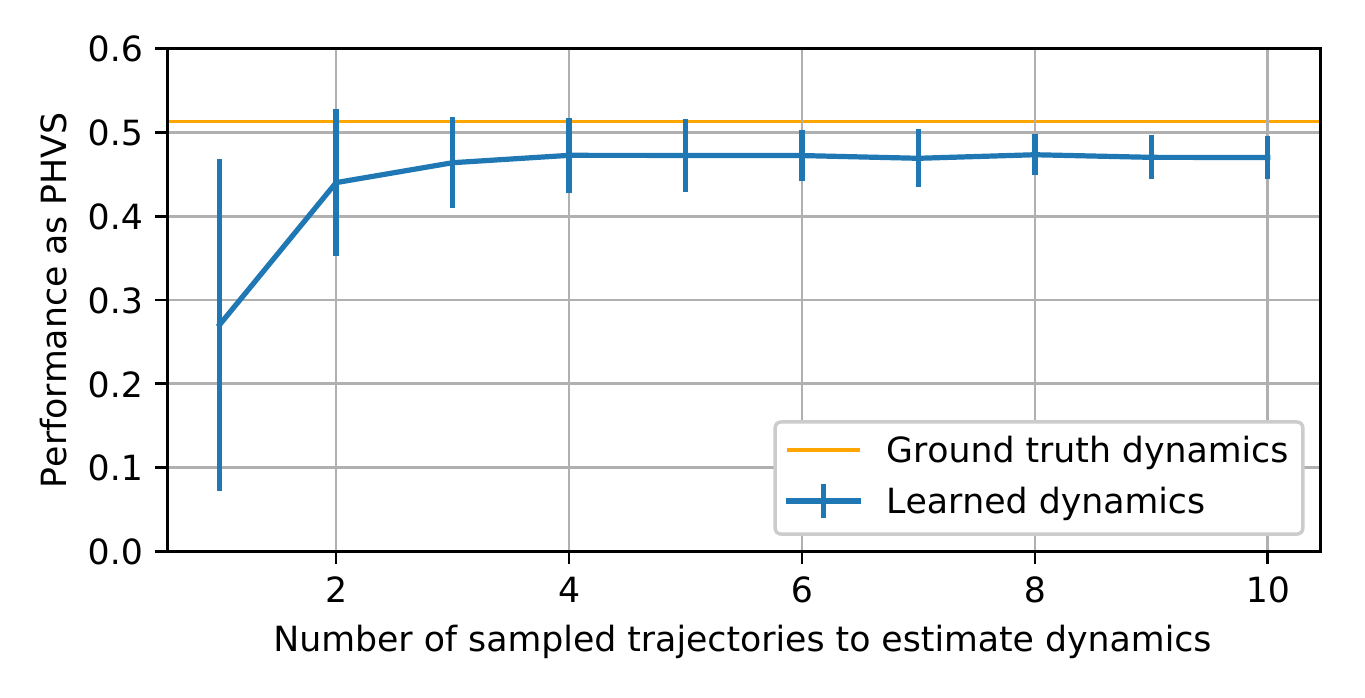}
    \caption{Performance comparison of iLQR optimized policies with different numbers of sampled trajectories for dynamics learning. The plot shows the mean value and one standard deviation for 500 optimization runs in simulation. Performance with ground truth dynamics is shown as a reference.}
    \label{fig:result_dyn_learning}
    \vspace{-0.6cm}
\end{figure}
\begin{figure*}[h]
    \centering
        \includegraphics[width=1.0\linewidth,clip,trim={0 15 0 15}]{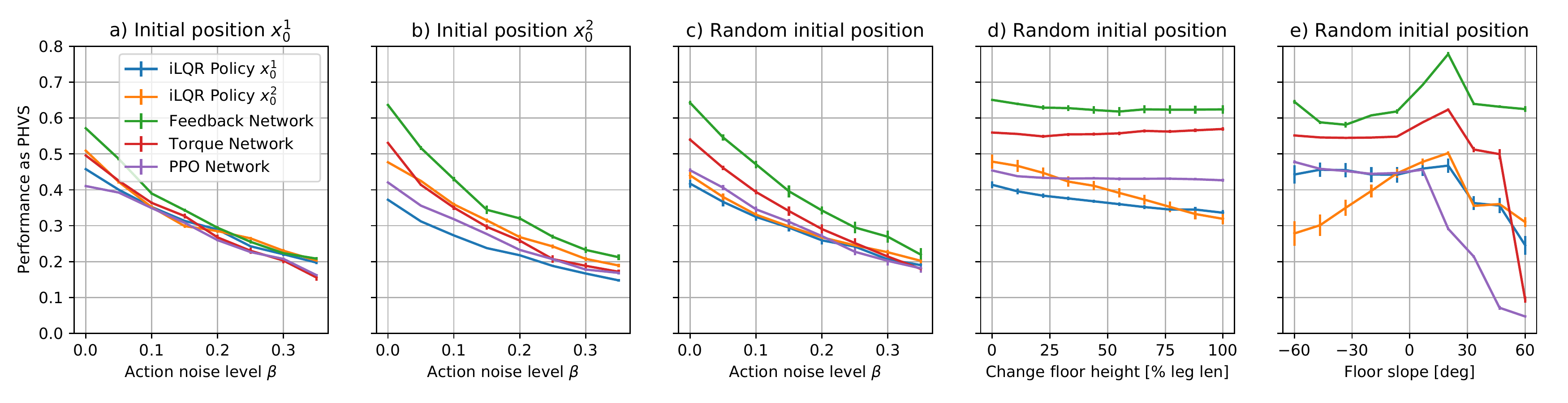}
    \caption{a) and b) Performance for various levels of action noise when starting from the trained initial positions $x_0^1$ and $x_0^2$ making 10 rollouts with the same policy. c) Average performance over 100 random initial configurations for increasing action noise levels. d) and e): Average performance of 100 random initial configurations for increasing floor height and various slopes. }
    \label{fig:result_noise}
    \vspace{-0.7cm}
\end{figure*}
\label{sec:results}
In this section, we present and discuss the experimental results. The goal of our experiments it to analyze each element of the proposed algorithm, in particular: 1) the efficiency of our dynamics regularization to learn local models, 2) the performance of the learned policy and its generalization capabilities, 3) the structure of the feedback network policy, 4) performance on real hardware. When possible, we also compare our results with the PPO Network Policy. The total runtime on a MacBook Pro with 2.9 Ghz Intel Core i7 for optimizing the local policies, learning the dynamics and training the torque and feedback network is 5 minutes. On the same system, the PPO reward converges after 5:40 minutes.
\subsection{Dynamics regularization}
We aim to analyze the effect of the dynamics regularization and in particular how it helps to reduce the numbers required trajectories to fit the dynamics model. For this, we optimize iLQR policies in simulation for a trajectory of \SI{0.5}{\second} over 30 optimization iterations from the $x^1_0$ initial position. We increase the number of sampled trajectories to estimate the dynamics from 1 to 10, always using a GMM prior with 5 mixture components. The sample trajectories are created by running the current iLQR policy with $\b{\Sigma}_t=0.01\mathds{1}$ in \cref{eq:ilqr_policy} from $x^1_0$ on the system. We compute statistics over 500 runs for each configuration on the PHVS metric.

Without the prior we were not able to fit a good dynamic model to drive the iLQR optimization even when using up to $1000$ samples. Therefore, \cref{fig:result_dyn_learning} shows results only with the dynamics regularization.

As one can see, the PHVS performance increases till 5 sampled trajectories per optimization iteration. Assuming the increase in the PHVS performance is due to better estimated dynamics for the optimization, this also implies the estimated dynamics quality improves up to 5 samples. The achieved optimization performance is quite close to the one when using the ground truth dynamics provided by the simulator. Increasing the number of samples further has no significant effect anymore. Note that until 3 samples, the standard deviation in the performance is quite high. This indicates that the resulting trajectories are quite different in terms of performance for a lower number of dynamics samples and the optimization results vary a lot.
{\setlength{\tabcolsep}{0.5em} 
\renewcommand{\arraystretch}{1.2}
\begin{table}[h]
\begin{center}
\begin{subtable}[t]{1.0\linewidth}
\begin{tabular}{|c c|c|c|c|c|}
\cline{3-6}
\multicolumn{2}{c|}{~} & \multicolumn{4}{c|}{Max slope} \\
\multicolumn{2}{c|}{~} & 15 deg & 30 deg & 45 deg & 60   deg \\
\hline
\multirow{4}{*}{\rotatebox[origin=c]{90}{Max height}}&  25\% & 9.4 $\pm$ 2.0 s & 8.4 $\pm$ 2.7 s & 8.3 $\pm$ 2.7 s & 6.0 $\pm$ 3.8 s\\
\cline{2-6}
& 50\% & 9.6 $\pm$ 1.8 s & 9.4 $\pm$ 1.8 s & 8.2 $\pm$ 2.9 s & 7.5 $\pm$ 3.3 s\\
\cline{2-6}
& 75\% & 9.6 $\pm$ 0.2 s & 8.5 $\pm$ 3.2 s & 8.0 $\pm$ 5.8 s & 6.4 $\pm$ 9.7 s\\
\cline{2-6}
& 100\% & 9.6 $\pm$  1.4 s & 8.9 $\pm$  2.6 s & 8.2 $\pm$  2.8 s & 6.6 $\pm$  3.3 s\\
\hline
\end{tabular}
\caption{Results for feedback network policy.}\label{table:res_marathon_feedback}
\end{subtable}
\vspace*{0.2cm}
\begin{subtable}[t]{1.0\linewidth}
\begin{tabular}{|c c|c|c|c|c|}
\cline{3-6}
\multicolumn{2}{c|}{~} & \multicolumn{4}{c|}{Max slope} \\
\multicolumn{2}{c|}{~} & 15 deg & 30 deg & 45 deg & 60   deg \\
\hline
\multirow{4}{*}{\rotatebox[origin=c]{90}{Max height}}&
25\% & 10.0 $\pm$ 0.0 s & 6.5 $\pm$ 3.4 s & 2.6 $\pm$ 2.3 s & 1.7 $\pm$ 1.4 s\\
\cline{2-6}
& 50\% & 10.0 $\pm$ 0.0 s & 6.5 $\pm$ 3.0 s & 3.0 $\pm$ 1.9 s & 1.8 $\pm$ 1.4 s\\
\cline{2-6}
& 75\% & 10.0 $\pm$ 0.0 s & 6.9 $\pm$ 3.0 s & 2.3 $\pm$ 1.6 s & 1.9 $\pm$ 1.7 s\\
\cline{2-6}
& 100\% & 10.0 $\pm$ 0.0 s & 6.3 $\pm$ 3.3 s & 2.1 $\pm$  1.8 s & 1.9 $\pm$  1.8 s\\
\hline
\end{tabular}
\caption{Results for PPO network policy.}\label{table:res_marathon_ppo}
\end{subtable}
\end{center}
\vspace{-0.5cm}
\caption{Termination time (mean and standard deviation) for a 10 seconds run with the feedback and PPO network policies over 50 runs. Max height as percentage of leg length.}
\label{table:res_marathon}
\vspace{-0.3cm}
\end{table}}
\vspace{-0.3cm}
\subsection{Comparison of policies on trained initial states}
\label{sec:result_sim_perf}
As a basic validation test for the learned neural network policies, we compare their performance to the iLQR policies in simulation. For this, we perform a rollout of the network policies from each of the iLQR policies' starting positions. To quantify the robustness of the policies, we set $\b{\Sigma}_t$ when sampling an action for the policies to $\b{\Sigma}_t=\beta \mathds{1}$. This has a similar effect as adding noise on the policy's action and we call this \italic{action noise}. We perform rollouts with $\beta~\in~[0.0, 0.35]$ and compute the PHVS metric. All rollouts are made for a trajectory of length \SI{1}{\second}.

The results are shown in the two left plots of \cref{fig:result_noise}. As we can see, the performance of all the policies declines as the action noise increases. In particular, the torque and feedback network policies have similar performance as the iLQR policies. For the second initial position, the feedback network policy performs significantly better than the iLQR policies. While not as good as the feedback network policy, the torque network policy also performs better than the iLQR policies.
This is interesting, as both network policies are trained on the data from the iLQR policies but they lead to higher performance, especially as noise increase. Overall, the PPO Network Policy behaves similarly to the other network policies, though usually slightly worse.

The feedback and torque neural network policies can reproduce at least
the same performance as the iLQR policies and in certain cases they outperform those policies despite being optimized on the same initial conditions. PPO never performs better than the other network policies.
\vspace{-0.3cm}
\subsection{Generalization}
Next, we study the generalization capabilities of the learned policies.
First, we test the policies for random initial configurations on a flat ground.
Then we choose random starting position and change either the ground height (decreased at $t=\SI{0.3}{\second}$ between $[0\%, 100\%]$ of the robot's leg length) or
the slope (between $[-60,60]$ degrees). We perform each test for 100 different initial configurations. The robustness of the policies is determined by measuring the PHVS for different level of action noise as in \cref{sec:result_sim_perf}.

Figure 5c) shows the PHVS performance for various initial
starting positions. As expected, the iLQR policies perform the worst.
In contrast, the performance of the network policies is similar to the first two starting positions and sometimes even higher. These results are not surprising: The iLQR policies are trajectory-centric
in the sense that they are policies around a single trajectory. Starting their policy from a different initial state easily causes them to become unstable. In contrast, the network policies depend only on the current state and have no notion of time or initial position and can therefore react to a new changing starting position. The PPO policy shows similar results although always performing slightly worse.

When changing the ground height, the network policies perform very well for larger changes. While the PPO policy keeps a similar performance over various height changes it has a worse PHVS measure than the other policies. In the case of slope changes,
the jump often fails when the robot knee or hip enters in collision with the ground. In this case both PPO and torque policies degrade when the slope is too high.
Overall the feedback network policy perform best.
Finally, we test the policies in constantly changing environments.
We run them for \SI{10}{\second} while randomly changing the floor height and slope. More precisely, we let the neural network policy control the robot until the robot falls or hits the maximum rollout length of \SI{10}{\second}. Whenever the hopper is at an apex point, the floor height and floor slope are randomly changed. The new floor height and floor slope are sampled uniformly from $[-\text{maxFloorHeight}, \text{maxFloorHeight}]$ and $[-\text{maxFloorSlope}, \text{maxFloorSlope}]$. If the new floor configuration is in collision with the robot, new height and slopes are sampled. We compute the mean and standard deviation of the termination time over 50 rollouts.

The results for this experiment are shown in \cref{table:res_marathon}. We omit the results for the torque network policy as the results are qualitatively similar to the feedback policy.
Looking over all results in \cref{table:res_marathon_feedback}, the average terminal time of the feedback neural network policy stays roughly the same when the changes to the floor height increase. However, when the floor slope increases, the trajectory terminates earlier. In addition, the behavior can generalize to random changes in the floor slope and height.
The performance of the PPO network policy as shown in~\cref{table:res_marathon_ppo} is also more or less constant when changing the floor height. When changing the floor slope, the performance deteriorates quickly compared to the feedback neural network policy.
{\setlength{\tabcolsep}{0.5em} 
\renewcommand{\arraystretch}{1.2}
\begin{table}[h]
\vspace{-0.2cm}
\begin{center}

\begin{tabular}{|c c|c|c|c|c|}
\cline{3-5}
\multicolumn{2}{c|}{~} & \multicolumn{3}{c|}{Network Policy} \\
\multicolumn{2}{c|}{~} & Torque & Feedback & PPO \\
\hline
\multirow{4}{*}{\rotatebox[origin=c]{90}{Slope }}&
$25.0^{\circ}$ & 0.076 $\pm$ 0.017 & 0.128 $\pm$ 0.008 & 0.155 $\pm$ 0.004\\
\cline{2-5}
& $12.5^{\circ}$ & 0.128 $\pm$ 0.011 & 0.162 $\pm$ 0.004 & 0.217 $\pm$ 0.009\\
\cline{2-5}
& $0.0^{\circ}$ & 0.141 $\pm$ 0.007 & 0.182 $\pm$ 0.008 & 0.213 $\pm$ 0.008\\
\cline{2-5}
& $-12.5^{\circ}$ & 0.098 $\pm$ 0.008 & 0.125 $\pm$ 0.018 & 0.133 $\pm$ 0.011\\
\cline{2-5}
& moving slope & 0.097 $\pm$ 0.008 & 0.121 $\pm$ 0.020 & 0.167 $\pm$ 0.025\\
\hline
\end{tabular}
\vspace{-0.2cm}
\end{center}
\caption{Results on real hardware for \SI{2}{\second} runs (PHSV mean and one standard deviation computed over 5 runs).}
\label{table:res_real_policy_phvs}
\vspace{-0.2cm}
\end{table}}
\vspace{-0.3cm}
\subsection{Interpreting the learned control structure}
\begin{figure}
    \centering
%
      \includegraphics[width=1.0\linewidth,clip,trim={0 15 0 25}]{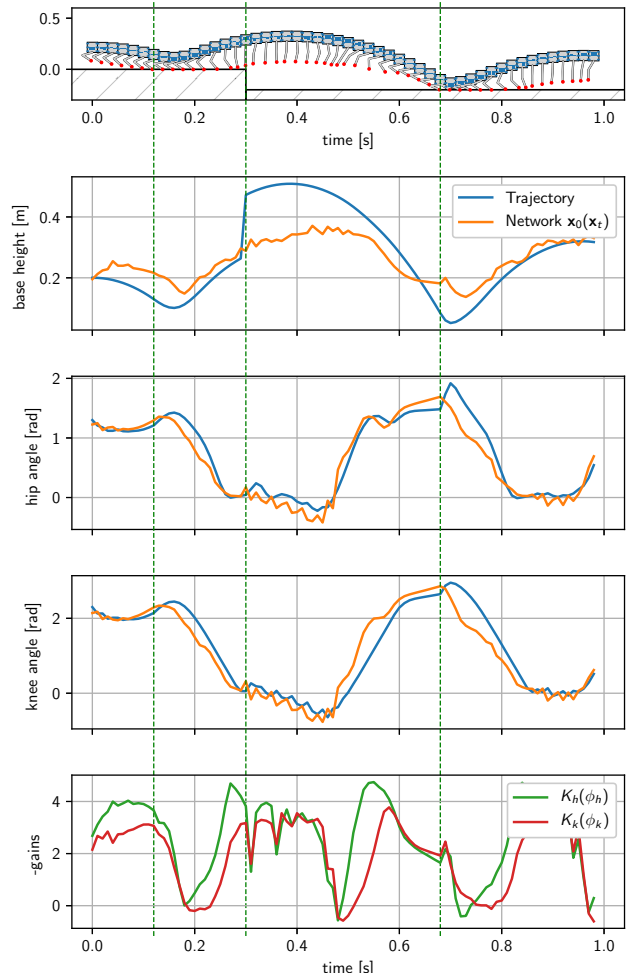}
%
    \caption{Top plot: snapshots of the robot over time. Middle plots: desired trajectories output from the feedback network policy and corresponding simulated trajectories. Bottom plot: diagonal elements of the gain matrix output by the policy.}
    \label{fig:result_nominal_traj}
\vspace{-0.6cm}
\end{figure}
%
One of the goal of this work is to learn a control policy with a structure that can
be interpreted.
In order to check if such structure was indeed learned, we run the network from starting position $x^1_0$ for a length of \SI{1}{\second} and change the floor height by one leg length at $t=\SI{0.3}{\second}$ in simulation. We then analyze at the output of the network, in particular to the nominal trajectory and the gains outputs (\cref{fig:result_nominal_traj}).

We compare the simulated robot motion with the desired nominal trajectory output from the feedback network policy, i.e. for every state $\xt$ along the trajectory we compute $\hat{\b{x}}^\text{feedback}_\theta(\b{x}_t)$.
First we notice that the robot closely follows (with a slight delay) the desired joint position trajectories
output by the network. This suggests that the network correctly encoded desired joint positions.

When the floor height changes at $t=\SI{0.3}{\second}$ the desired base height from the feedback network policy does not increase rapidly: it asks the robot to come down (note that the base height while jumping is uncontrollable).
Around $t=\SI{0.7}{\second}$, the robot gets close to the ground. Here the desired base position increases, the robot should start to come up and after $t=\SI{0.7}{\second}$ smaller desired angles for the hip and knee joint. These are reasonable desired positions which would make the robot jump when properly tracked.

The bottom graph of \cref{fig:result_nominal_traj} shows the time evolution of gains that are responsible for producing hip (knee) torques from the hip (knee)
position error. At impact, these gains decrease very quickly, leading to a more compliant behavior to absorb the impact. Then the joints stiffen to allow the robot to jump. The gains remain stiff during the flying phase, allowing the robot to reposition its leg, until the next impact. Interestingly, at the moment where the next impact would have been if the ground position would not have changed (at $\simeq 0.5$s) the gains again decrease, which
suggests that the policy anticipated a contact and increased compliance.

Overall, the gain schedule is consistent with the jumping
motion (stiff to jump, compliant at landing) and the desired positions correspond to the actual motion of the robot. Note that, while this can seem obvious, it was not clear that these outputs would have any meaning because network outputs are redundant and the actual torque sent to the robot is a combination of the outputs of the policy network.
These results suggest that the learned network outputs have an interpretable meaning and can help better understand what the policy tries to achieve at different instants of time.
This is an interesting aspect as usually neural network torque policies do not
have a structure that afford an understanding of the behavior in terms of desired positions and leg stiffness.

%
\subsection{Transfer to real hardware}
Finally, we test the transfer of the learned policies on the real robot.
We compute the PHVS metric for each policy over five executions of $t=\SI{2.0}{\second}$ on the real robot
for flat ground and grounds with slopes between $-12.5^{\circ}$ and $25.0^{\circ}$ (maximum slopes before robot started to slip from initial configuration).
We also tested the policies when the slope and height of the ground was continuously changing.
As can be seen on the attached video, the right side of the ground is moved up and down by a motor at a constant frequency while the left side is fixed. The slope changes between $-5$ and $15$ degrees.

Result are shown in \cref{table:res_real_policy_phvs}.
For all experiments, the three policies never failed. The PPO policy has a slightly higher PHSV than
the network policy. When analyzing the results, we find that the robot jumps at the same height for both policies. The PPO policy tends to keep the leg more to the back and is more compliant which makes the robot come closer to the ground at contact and therefore needs a higher upward velocity, which seems to explain the difference in PHSV.
The torque policy systematically jumps lower than both PPO and feedback policies and has a much lower PHSV.

Interestingly, the dynamics of the simulated robot is quite different from the real robot. First, we underestimated
the mass distribution of the robot as can be seen in  \cref{fig:hopper_setup}. For the simulation,
we used the CAD model of the robot but forgot to take the electronics and sensors into account. Additionally, the simple contact model used in the simulation was sufficient to transfer the behaviors on the real robot where there are non trivial friction effects and slipping.
Still, the reactive feedback policies can be
transferred on the real robot and are robust to a changing environment.
We were also able to transfer policies learnt with PPO but it is important to note that the policies in this case are not interpretable and it takes an order of magnitude more samples to train.
\section{Discussion and Conclusion}
\label{sec:conclusion_and_future_work}
In this paper, we considered the problem of learning control policies for dynamic tasks with significant impact
dynamics. We extended previous methods to work with such systems by 1) proposing a regularization method to learn a sample efficient dynamics model with switching contacts for underactuated systems,
2) describing a modified iLQR algorithm with adaptive receding horizon length that take into account actuation torques
and 3) learning time-independent, reactive policies with two types of neural networks, and in particular with a network that preserves the structure of the LQR policy, making the network amenable to analysis.

As our results show, we are able to learn a dynamics model with only 5 sampled trajectories per iteration. The trained network policies perform better on the training scenarios and new validation scenarios than the iLQR policies. The feedback network policy performs better, though not significantly better than a network that learns to predict actions directly. However, the feedback network policy is capable in providing more interpretable control outputs.  We envision that such policies could be more easily analyzed with formal control methods, afford a physical interpretation of the behavior (e.g. in terms of impedance regulation)
and be easier to reuse across different tasks (e.g. the feedback gains). It will be the focus of our future work. Finally, experiments on the real robot demonstrate that the learned policies can be effectively transferred on a real hardware while retaining their generalization properties.

When comparing our method to PPO, a state of the art reinforcement learning method, we found the feedback network policy to always have similar or better performance than PPO, while using one order of magnitude fewer samples during training. When changing the ground in simulation, our method tended to outperforms the PPO network policy but both PPO and the feedback network policy showed good performance on the real robot.
It is likely that PPO would perform better if it were trained under a larger variety of environments. It would then be interesting to compare our approach and PPO on a larger set of environments to study their generalization capabilities further.
%


\addtolength{\textheight}{5cm}   


\bibliographystyle{IEEEtran}
\bibliography{refs}{}

\end{document}